\documentclass[letterpaper, 10 pt, conference]{ieeeconf}  
\usepackage{amssymb}
\usepackage[dvipsnames]{xcolor} 
\usepackage{amsmath}
\usepackage{cancel}
\usepackage{graphics} 
\usepackage{epsfig} 
\usepackage{mathptmx} 
\usepackage{times} 
\usepackage{amsmath} 
\usepackage{amssymb}  
\usepackage{comment}

\usepackage{algorithmic}
\usepackage{algorithm}

\usepackage{lineno}

\IEEEoverridecommandlockouts                              


\newcommand{\DP}[1]{{\color{ProcessBlue} #1}}
\newcommand{\ST}[1]{{\color{Red} #1}}
\newcommand{\DPdelete}[1]{{\color{Melon} #1}}

\usepackage{stackrel,amssymb}
\newcommand{\leftrarrows}{\mathrel{\raise.75ex\hbox{\oalign{%
  $\scriptstyle\leftarrow$\cr
  \vrule width0pt height.5ex$\hfil\scriptstyle\relbar$\cr}}}}
\newcommand{\lrightarrows}{\mathrel{\raise.75ex\hbox{\oalign{%
  $\scriptstyle\relbar$\hfil\cr
  $\scriptstyle\vrule width0pt height.5ex\smash\rightarrow$\cr}}}}
\newcommand{\Rrelbar}{\mathrel{\raise.75ex\hbox{\oalign{%
  $\scriptstyle\relbar$\cr
  \vrule width0pt height.5ex$\scriptstyle\relbar$}}}}
\newcommand{\longleftrightarrows}{\leftrarrows\joinrel\Rrelbar\joinrel\lrightarrows}

\title{\LARGE \bf
Dimensionality Reduction of Dynamics on Lie Manifolds via Structure-Aware Canonical Correlation Analysis
}

\author{Wooyoung Chung$^{1}$, Daniel Polani$^{2}$, Stas Tiomkin$^{1,*}$\\{\small $^{1}$Charles W. Davidson College Of Engineering,
San Jose State University, CA, USA}\\{\small$^{2}$School of Physics, Engineering and Computer Science, University of Hertfordshire, UK}
\\{\hspace{0.7cm}\small wooyoung.chung@sjsu.edu, d.polani@herts.ac.uk, stas.tiomkin@sjsu.edu}
\thanks{$^*$ Corresponding Author}
}

\makeatletter
\let\NAT@parse\undefined
\makeatother
\usepackage{hyperref}

\begin{document}

\pagenumbering{arabic}

\maketitle
\thispagestyle{empty}
\pagestyle{empty}

\begin{abstract}
    Incorporating prior knowledge into a data-driven modeling problem can drastically improve performance, reliability, and generalization outside of the training sample. The stronger the structural properties, the more effective these improvements become. Manifolds are a powerful nonlinear generalization of Euclidean space for modeling finite dimensions. Structural impositions in constrained systems increase when applying group structure, converting them into Lie manifolds. The range of their applications is very wide and includes the important case of robotic tasks. 
    Canonical Correlation Analysis (CCA) can construct a hierarchical sequence of maximal correlations of up to two paired data sets in these Euclidean spaces. We present a method to generalize this concept to Lie Manifolds and demonstrate its efficacy through the substantial improvements it achieves in making structure-consistent predictions about changes in the state of a robotic hand.

\end{abstract}

\section{Introduction}

{Simplicity that respects structure is a desirable property of effective control. Typically, simplicity improves a method's robustness, feasibility, and flexibility and is achieved by reducing its complexity. A common form of complexity reduction is realized by dimensionality reduction, which represents a special case of the more general principle of information compression methods. 

One popular information compression model is the Information Bottleneck (IB) \cite{tishby2000information, slonim2002information}. A principled method to achieve such reductions and intimate links to statistical machine learning. 

Of particular interest to the control community is the fact that the IB is a direct generalization of the well-established CCA \cite{katayama2005subspace}. In the case of (locally) linear Gaussian models, CCA permits tuning the degree of structural preservation from one variable to another. The IB  thus implements a "soft" CCA in the Gaussian case \cite{chechik2003information, creutzig2009past}.

 However, CCA and its informational generalization (IB) purely concentrate on preserving the dependency of the target variables. They are utterly indifferent to any particular additional structure of the problem, some paradigmatic consequences of which we now illustrate in a pertinent example.

In \cite{creutzig2009past}, an IB is applied to a linear Gaussian control channel which thus reduces to a soft CCA model. Reducing the information that a simpler model has access to the process leads to a progressive reduction of the dimensionality in the accompanying soft CCA. However, this reduction is purely correlational and does not consider the special structure of the control loop, as discussed in \cite{amir2015past}. Concretely, in the controlled system, the matrix transforms the process 'past dynamics into the process' future dynamics. The original transformation has a particular recursive structure following that of the Hankel matrix \cite{katayama2005subspace}. After the system is naively subject to information/dimensionality reduction, the resulting reduced transformation matrix for the compressed system no longer has a Hankel structure \cite{creutzig2009past}. 

The structural deficit is alleviated in \cite{amir2015past} by modifying the approach of \cite{creutzig2009past} to constrain the reduced transformation matrices to satisfy the properties of a proper Hankel matrix to represent actual control systems. However, this approach is unsuitable for general use due to the requirement of handcrafting the information reduction method to respect the Hankel matrix structure of the control problem.

This work presents a method to support a generalizable approach to produce structure-respecting  IB in the future. We consider the case where a modification to the traditional CCA method respects a constrained structure, specifically, a manifold to which the variables of interest and its interrelation are confined. In short, we present a method where we constrain our problem space to manifolds instead of Euclidean space.

To achieve this, we note that the naive concept of various averaging operations available in the Euclidean case needs to be modified, as manifolds do not offer mean and variance computation via vector addition operations. Therefore, instead, we resort to the variational description of the quantities of interest. The idea is analog to the fact that, in Euclidean space, the centroid of given data points can not only be computed by taking the average vector of the data points but, alternatively, by minimizing the sum of the squared Euclidean distances. 

The proposed method's key components will be manifold-intrinsic distance, replacing the averaging operation to find the mean by variational computations, and the projections to sub-manifolds instead of Euclidean vector subspaces.

}

{
{The paper is organized as follows. Section~\ref{background} begins with a general overview of well-known concepts and literature reviews.} An overview is given of the formalism in Section~\ref{overview} followed by the preliminaries relevant to the development of the formalism in Section~\ref{sec:prelim} (note that Lie Theory and manifolds as relevant to the paper are covered in the Appendix~\ref{sec1:firstappendix}). We highlight how to replace the computation of the mean and centroid with a variation of expected distance to transfer the concept from Euclidean space onto Lie manifolds. This step is crucial for the development of the Intrinsic CCA. We then discuss the Intrinsic PCA's main features, which are useful for building the intuition for its extensions towards the ICCA. Section \ref{sec:method} presents the proposed method for ICCA,  extending the concept of principal geodesic curves
to canonical geodesic curve pairs, denoted by {\it intrinsic CCA}, and represents an efficient algorithm for the calculation of ICCA from data points on high-dimensional Lie manifolds. In Section
\ref{sec:Experiments}, we demonstrate this algorithm on a high-dimensional articulated robotic system, an anthropomorphic robotic hand, whose configuration space is given, in general, by the corresponding multi-dimensional Lie manifold.
Section \ref{sec:conclusion} reanalyzes the paper to understand the impact of the findings and build potential future directions\footnote{All the experiments and results can be reproduced by our code repository: \url{https://github.com/JWK7/ICCA}}.
}

\section{Overview}
\label{background} \label{overview} 
Existing methods for dimensionality reduction consider their source data living in "flat" Euclidean spaces and are utterly agnostic to any potential additional structure or constraints \cite{katayama2005subspace, chechik2003information, klami2013bayesian, andrew2013deep,wang2015stochastic}. Specifically, there are no methods for \emph{joint} dimensionality reduction (compression) of two sets of points when these are restricted to a Lie manifold. 

Joint compression could reveal predictive models between one set to another, with additionally choosing a desired level of complexity and details. In particular, such structure-preserving predictive models would permit the reconstruction of a point in one set on the Lie manifold from a point in another set on the same Lie manifold. A particular scenario of interest is the estimation of dynamical models on the manifold, where two sets of points on the manifold represent the current state and the state in $T$ time steps to the future. This estimation is impossible if a model does not preserve the manifold structure.





Here, propose a concrete method to generalize the CCA to the ICCA on general Lie manifolds.
Amongst others, the Riemannian approach provides a means to account for additional intrinsic structure and control a model's intrinsic complexity by choosing how many intrinsic geodesic pairs are being used (Section~\ref{sec:method}). This is achieved by using projections and optimization compared to the conventional correlational matrix approach. This increases the accuracy and flexibility of the mapping made through the canonical pairs compared to traditional (extrinsic) CCA.

We demonstrate ICCA on a simulated robotic hand with the state space represented on the Lie Manifold. These high-dimensional dynamical systems have been used extensively in control research\cite{88036,Ahmed2015Sensor,5460699}. Our main contributions include:
\begin{enumerate}
    \item Two algorithms for ICCA decomposition and reconstruction.
    \item Reduction in the prediction error of the future state of robotic hand from the current state compared to the existing baseline dimensionality reduction methods;
    \item Structural guarantees that the predicted state is confined to the Lie manifold;
\end{enumerate}
As an unplanned outcome, we found that the relationship between the "intrinsic" times of the basic geodesic movements can be mapped linearly to each other (Section \ref{sec:Experiments}).

\vspace{-0.15cm}
\section*{Notation}
Scalars are denoted by the lowercase letters, e.g., $t\in \mathbb{R}^+$. A Lie group and its corresponding algebra are denoted by $G$ and $\mathfrak{g}$, respectively. For special groups, such as $SO(3)$, we write the algebra as its lowercase pendant, such as $so(3)$. The Euclidean (extrinsic) distance between two points $x_1$ and $x_2$ is given by $||x_2-x_1||_2$, while the Riemannian (intrinsic) distance between such points on a manifold is denoted by $D(x_1,x_2)$.

\section{Preliminaries}\label{sec:prelim}

This work uses the standard definitions of the Lie manifold (including the groups for rotation and translation) frequently used in robotics. This background is provided in Appendix~\ref{sec1:firstappendix} for completeness. 

The key components of the proposed method (intrinsic distance, averaging operations, and sub-manifold projection) are over-viewed in the next section. 

\subsection{Intrinsic vs.\ Extrinsic Means}\label{sec: Intrinsic vs Extrinsic Means}

Intrinsic and extrinsic calculations differ fundamentally from one another. Extrinsic (Euclidean) calculation limits itself to the linear constraint and does not account for the non-linear structure embedded in many data. To develop the structure-aware ICCA, we will need to calculate the intrinsic mean and projections of data points on the Lie manifold to its sub-manifolds (Section \ref{sec:Projection to Subgroups}). 



We now discuss the necessary background to define these concepts.

Given a set of data points $\{x_i\}_{i=1}^N$ in a metric space, $\mathcal{X}$, 
their mean, $\mu_{x}$, is defined by:
\begin{align}
  \label{eq:mean}
  \mu = \underset{x\in \mathcal{X}}{\mbox{ argmin }} \sum_{i=1}^N D^2(x, x_i),
\end{align}
where $D(\cdot, \cdot)$ denotes the distance between points in
$\mathcal{X}$ and is assumed the minimum to be unique. Thus, the mean of a set of points in a general metric space is given by the solution to the optimization problem in Eq.~\eqref{eq:mean}.

This variational formulation offers a generalization of the mean in a space in which an arithmetic mean can be computed, i.e.\, which permits convex combinations or explicit addition operators, such as the standard Euclidean space. In the latter, Eq.~\eqref{eq:mean} can be solved in closed
form, namely as the arithmetic average of the data points,
$\mu_x=\frac{1}{N}\sum_i x_i$. 

\subsubsection{Intrinsic mean} 
 
We now apply this variational method to compute the mean intrinsically to a Lie manifold (Appendix~\ref{sec1:firstappendix}). Given a Lie manifold, $\mathcal{X}$, the distance between data points, $x_1, x_2\in \mathcal{X}$ is given by the Riemannian distance on the manifold: 
\begin{align}
D^2(x_1, x_2) \triangleq ||\log (x_1^{-1}x_2)||^2_2\label{eq:Riemannian distance},
\end{align}
where the inverse is interpreted in the sense of the Lie group
inverse. With this, we generalize the calculation of the intrinsic
mean \cite{fletcher2003gaussian, moakher2002means} of $x_1, x_2, \dots, x_N\in \mathcal{X}$ by solving:
\begin{align}
  \label{eq:intrinsic-mean}
\mu = \underset{x\in \mathcal{X}}{\mbox{ argmin }} \sum_{i=1}^N D^2(x_1, x_2).
\end{align}
The problem in Eq.~\eqref{eq:intrinsic-mean} can be solved iteratively \cite{fletcher2003gaussian, moakher2002means} or can be approximated by the Baker–Campbell–Hausdorff formula \cite{bonfiglioli2011topics} given by Eq.~\eqref{eq:CHB}:
\begin{align}
    ||\log (x_1^{-1}x_2)||_2 \approx ||\log (x_2) - \log (x_1)||_2\label{eq:CHB},
\end{align}
which omits the non-commutative terms between $x_1^{-1}$ and $x_2$. 

\subsubsection{Extrinsic mean} If we instead embed data points from the Lie
manifold $\mathcal{X}$ into an ambient Euclidean space
\cite{agrachev2013control} we can calculate the mean in
Eq.~\eqref{eq:mean} directly, using the Euclidean distance of the ambient space and the arithmetic computation of the mean,
which leads to the conventional {\it extrinsic mean}. However, in general, the extrinsic
mean will not be a point on a manifold. This creates a discrepancy between
the internal structure of the data and its {\it extrinsic} statistics.
This discrepancy results in imprecise modeling of data if these are actually located on a manifold
and, hence, in deficient generalization across samples.
{Even worse, since it violates the constraints represented by the manifold structure, in these cases the extrinsic mean
may not even represent a physically realizable configuration of the system at all.}

\subsection{Projection to Subgroups}\label{sec:Projection to Subgroups}
Let ${G}$ and $\mathfrak{g}$ be the Lie group and its corresponding algebra. 
For an arbitrary unit vector ${v}\in \mathfrak{g}$, we can define a one-parameter subgroup $H_{v}$ of ${G}$ \cite{fletcher2003gaussian}:
\begin{align}
    H_v \triangleq \bigl\{\exp (t{v})\in G\;:\; t\in \mathbb{R}\bigr\},
\end{align}
where '$\exp$' is the exponential map from $\mathfrak{g}$ to ${G}$, given in Eq.\eqref{eq:expmap}.
The distance between any ${x}\in G$ and $H_v$ is given by:
\begin{align}
D({x}, H_{v}) \triangleq&\;\; \underset{t}{\mbox{min}} \; D({ x}, \exp(t{v})),\label{eq:distg_Hv}\\
\intertext{with the optimal value of $t$ being given by:} 
t^*=&\underset{t}{\mbox{ argmin}} \; D({ x}, \exp(t{v})),\label{eq:optT}
\end{align}
determining the {\it projection} of ${x}$ onto $H_v$:
\begin{align}
    \mbox{Proj}_{H_{v}}({ x})\triangleq\exp(t^*{v}).\label{eq:proj}
\end{align}
This projection of a group element to a one-parameter subgroup is a
core component in the Intrinsic PCA \cite{fletcher2003gaussian, fletcher2003statistics} (explained in the next section), which we generalize to Intrinsic CCA in this work.




{

\subsection{Intrinsic Principal Component Analysis}
To provide the intuition for the development and the key features in the proposed Intrinsic CCA method, we recapitulate an existing Intrinsic Principal Component Analysis (PCA) method on Lie manifolds \cite{fletcher2003statistics}. Traditional  PCA is concerned with dimensionality reduction through the projection of data to linear subspaces, minimizing the reconstruction error. As one increases the dimensionality of the subspaces, new independent (orthogonal) additional features are accounted for.

Whether in the Euclidean or the manifold case, the calculation is based on the above principle of hierarchical projection of data on those subspaces or -manifolds which minimize the mean projection error between the data points and the corresponding subspace \cite{fletcher2003statistics}. The difference between PCA in the Euclidean space and PCA on the Lie manifold is in the definition of the subspace and that of the distance, and the operators used for averaging 
(Appendices~\ref{sec: Intrinsic vs Extrinsic Means} and \ref{sec:Projection to Subgroups}).


We proceed by presenting first the calculation of PCA in both  Euclidean space and the Lie manifold in order to prepare the background to the introduction of the ICCA method.

\noindent{\bf PCA in Euclidean Space} One calculates the hierarchical projections beginning with the the first PCA component, $k=1$, which is a one-dimensional linear (strictly spoken, affine) space. Given the
Euclidean space, $X$, we compute it by seeking a one-dimensional subspace onto which the data points $x_1, x_2,
\dots, x_N\in X$ project with the least total distance loss, more precisely, we seek  $S_v=\{t{v}\;:\;
t\in \mathbb{R}\}$ such that \cite{fletcher2003statistics}: 
\begin{align}
    {v}^{(1)} &= \underset{||{v}||=1}{\mbox{ argmin }}\sum_{i=1}^N ||x_i -   \mbox{Proj}_{S_v}(x_i)||^2_2\label{eq:pca1},\\
    \intertext{where $\mbox{Proj}_{S_v}(x)=(x\cdot
{v}){v}$ is the optimal projection of $x$ on
$S_{{v}}$. We compute the subsequent PCA components recursively by proceeding to increasingly higher-dimensional subspaces, by removing the contribution of the already established subspaces and minimizing the distance loss of the data points with respect to the newly added one. Concretely, for $k>1$, we  calculate
 recursively \cite{fletcher2003statistics}:}
    {v}^{(k>1)} &= \underset{||{v}||=1}{\mbox{ argmin }}\sum_{i=1}^N ||x_i - \mbox{Proj}_{S_v}(x_i) - 
                         \sum_{\ell=1}^{k-1}
                         \mbox{Proj}_{S_{v^{(\ell)}}}(x_i)
                         ||^2_2. \label{eq:pca2}
\end{align}

In other words, the first projection minimizes the deviations to the first component, and all subsequent projections minimize the residual deviation to the new component after all previous components have been accounted for.

{There are two essential differences between PCA in the Euclidean space and on the Lie manifold. First, the Euclidean distance function is inappropriate for estimating the projection error on the Lie manifold. Instead, the manifold-intrinsic distance, such as the Riemannian distance, should be used \cite{PCALie}. Second, the sub-spaces/principal components in the Euclidean space are given by vector (strictly spoken, affine, if the mean does not coincide with the origin) subspaces, while in the Lie manifold, they are given by {\it principal geodesic curves}.}

\noindent{\bf PCA on Lie Manifolds} The generalization of the first principal vector in Eq.~\eqref{eq:pca1} to the first principal geodesic curve is achieved by combining Eq.~\eqref{eq:mean},
Eq.~\eqref{eq:distg_Hv}, and Eqs.~\eqref{eq:pca1} and \eqref{eq:pca2}. One obtains the principal geodesic curve, \cite{fletcher2003statistics}:
\begin{align}
    {v}^{(1)} = \underset{||{v}||=1}{\mbox{ arg{ min} }} \sum_{i=1}^N\underset{t}{\mbox{ min}}||\log\bigl((\mu^{-1}x_i)^{-1}\exp(t{v})\bigr)||^2,
\label{eq: IPCA}
\end{align}
where $\mu$, $'\log'$, and $'\exp'$, are the mean, given in
Eq.~\eqref{eq:mean}, the logarithmic and
exponential maps (cf., Appendix), accordingly. The principal geodesic curves for $k>1$ are defined analogously to ${v}^{(k>1)}$ in Eq.~\eqref{eq:pca1}, again, with the appropriate distance (Eq.~\eqref{eq:distg_Hv}), projection (Eq.~\eqref{eq:proj}), and mean (Eq.~\eqref{eq:mean}), respectively. 

Following this intuition and the properties of projections to a submanifold, we now extend the Intrinsic PCA method \cite{fletcher2003statistics} to the novel Intrinsic CCA method. Here sets of point pairs on the Lie manifold are recursively projected onto a sequence of pairs of submanifolds. In analogy to PCA, a current manifold pair recursively encompasses previous submanifold pairs to minimize the intrinsic error between the mapped points.

\section{Proposed Method - ICCA}\label{sec:method}

Canonical Correlation Analysis (CCA) is a fundamental tool to estimate two subspaces and a linear transformation between them from data, such that each of the original data sets are represented by (projected to) the corresponding subspaces with minimal error; similarly, there is a minimal error between the projected data to these subspaces.




Its applications extend far beyond pure data analysis; they include computer vision \cite{Kim2014Images}, speech synthesis \cite{Arora2012Sound}, and robotic control \cite{Lisanti2014Camera}. Notably, and of particular interest for control, CCA and its information-theoretic generalization has been proven useful for the dimensionality reduction of linear dynamical systems \cite{amir2015past, creutzig2009past, chechik2003information, katayama2005subspace}. With its wide variety of applications, however, conventional CCA tends to handle non-linear data poorly due to its Euclidean assumptions \cite{Chen2021Robot}. One can therefore expect that adapting CCA to the intrinsic structure of the data it is to represent should improve the ability of this dimension reduction technique to respect and faithfully preserve the mapping between the set of data pairs. 

In this work, we propose a particularly useful specialization of the nonlinear case, namely a CCA that "lives" specifically on Lie manifolds. Such an operation is of particular interest in the context of robotics \cite{LieRobotics}. Environments that contain complex manipulation, shape modifications, or other structural changes can be suitably described in the language of Lie group operations.}



Much like in PCA, CCA implements a hierarchy  of projections between data and sub-spaces. However, CCA uses two projections per component, called the \textit{canonical pairs}. In the Euclidean space these pairs are derived using the standard distance, averaging and projection operators. 


In order to define the intrinsic CCA, we thus translate its functionality, given in the first paragraph of this section, into the language of intrinsic distance, averaging and projection operators on the Lie manifold, as follows.




\subsubsection{Projection on the Subgroups} We define one-parameter subgroups $H_{{v}}$ and $H_{{u}}$ of $G$, a distance between $x,y\in G$ and $H_{{v}}$, $H_{{u}}$, respectively, 
where $v,u\in\mathfrak{g}$:
\begin{align}
H_{{v}} \triangleq \bigl\{\exp (t{v})\in G\;:\; t\in R\bigr\},\\
H_{{u}} \triangleq \bigl\{\exp (s{u})\in G\;:\; s\in R\bigr\},\\
D(x, H_{{v}}) \triangleq \underset{t}{\mbox{min}} \; D(x, \exp(t{v})),\label{eq:distX_Hv}\\
D(y, H_{{u}}) \triangleq \underset{s}{\mbox{min}} \; D(y, \exp(s{u}))\label{eq:distX_Hu},
\end{align}
{By using the base movement of (${v}$, ${u}$), we can represent each data point pair through the {\it optimal projection times} ($t^*$,$s^*$):}
\begin{align}
 t^*  =& \underset{t}{\mbox{ argmin}} \; D(x, \exp(t{v})), \\ s^*  =& \underset{s}{\mbox{ argmin}} \; D(x, \exp(s{u})).
\label{eq:optTime}
\end{align}
The distance, $D(\cdot, \cdot)$, in Eq.~\eqref{eq:distX_Hv} and Eq.~\eqref{eq:distX_Hu} is the intrinsic Riemannian distance Eq.~\eqref{eq:Riemannian distance}. 


\subsection{Intrinsic Canonical Correlation Analysis} 

Our methodology relies on optimizing projections to minimize the distance between the two subgroups of the set of points. We compress the original data into the pair (${v}, {u}$) with their corresponding projection times ($t^*,s^*$). (${v}, {u}$) and ($t^*,s^*$) are selected to minimize reconstruction errors from initial to final configuration data, corresponding to the first vs.\ second entry of the pair, respectively, as explained below.

\subsubsection{First ICCA pair}
Given $N$ point pairs, $\{x_i\in G, y_i\in G\}_{i=1}^N$, on the Lie
manifold, $G$, we defined the first canonical geodesic pair as a pair
of vectors in the corresponding Lie algebra, $\mathfrak{g}$,
$\bigl({v}^{(1)}\in \mathfrak{g}, {u}^{(1)}\in
\mathfrak{g}\bigr)$, representing two one-parameter subgroups
$\bigl(H_{{v}}, H_{{u}}\bigr)$, with which the projected data
$\bigl\{\mbox{Proj}_{H_{{v}}}(x_i)\in G,
\mbox{Proj}_{H_{{u}}}(y_i)\in G\bigr\}_{i=1}^N$ (with parametrizations
$\bigl\{t^*_i, s^*_i\bigr\}_{i=1}^N$) are maximally associated.

We propose to calculate the first ICCA pair from $N$ point pairs $\{x_i, y_i\}_{i=1}^N$ on the manifold by:
\begin{align}
{v}^{(1)}, {u}^{(1)} = \underset{\substack{||v||=||u||=1}}{\mbox{argmin}}\sum_{i=1}^N \Bigl(&D^2(\mu_x^{-1}x_i, H_{{v}})\label{eq:1stPair}\\
+&D^2(\mu_y^{-1}y_i, H_{{u}})\nonumber \\
+&D^2\bigl(\mbox{Proj}_{H_{{v}}}(\mu_x^{-1}x_i), \mbox{Proj}_{H_{{u}}}(\mu_y^{-1}y_i)\bigr) 
\Bigr).\nonumber 
\intertext{or, explicitly,}
{v}^{(1)}, {u}^{(1)} = \underset{\substack{||v||=||u||=1}}{\mbox{argmin}}\sum_{i=1}^N \Bigl(&\underset{t}{\mbox{ min }}D^2(\mu_x^{-1}x_i, \exp(t{v}))\label{eq:1stPair_eplicite}\\
+&\underset{s}{\mbox{ min }}D^2\bigl(\mu_y^{-1}y_i, \exp(s{u})\bigr)\nonumber \\
+&D^2\bigl(\exp(t_i^*{v}), \exp(s_i^*{u})\bigr) 
\Bigr),\nonumber 
\end{align}
where the first two terms define for each $i$ the time pair $ (t_i^*, s_i^*)$ via Eq.~\eqref{eq:optTime}, while the third term is the distance between the projected $x_i$ and $y_i$, with the corresponding $t_i^*$ and $s_i^*$, from each other. The joint projections given by the first two terms and the last term in Eq.~\eqref{eq:1stPair_eplicite} extend the intrinsic PCA towards the ICCA. 

\noindent{\bf Optimal Projection Time.} The solution to Eq.~\eqref{eq:1stPair} includes $\bigl({v}^{(1)}, {u}^{(1)}\bigr)$ and $\bigl\{t^*_i, s^*_i\bigr\}_{i=1}^N$. Given this solution we train a linear regression model, $f^{(1)}_{\psi}$, parameterized by $\psi$, for predicting $s$ from $t$: 
\begin{align}
\psi^* = \underset{\psi}{\mbox{ min }} \frac{1}{N}&\sum_{i=1}^N \bigl(s_i^* - f^{(1)}_{\psi}(t^*_i)\bigr)^2
\intertext{which we concisely denote by $\hat{s}(t)$:}
    \hat{s}(t) =& f^{(1)}_{\psi^*}(t). \label{eq:OptimalProjectionTime}
\end{align}
The model in Eq.\eqref{eq:OptimalProjectionTime}  allows us to reconstruct $\hat{y}$ from $x$ using the first ICCA pair by:
\begin{align}
t^{*} & = \underset{t}{\mbox{ argmin}} \; D(x, \exp(t{v}^{(1)})),\label{eq:recoTime}\\
\hat{y} &= \exp\bigl(\hat{s}(t^* ){u}^{(1)}\bigr)\label{eq:recoY},
\end{align}
where $D(\cdot, \cdot)$ and '$\exp$' are the (intrinsic) Riemannian
distance Eq.~\eqref{eq:Riemannian distance} and the exponential map 
Eq.~\eqref{eq:expmap}, respectively. 

\begin{figure}[t!]
\centering\hspace{-1.35cm}
\includegraphics[scale=0.45]{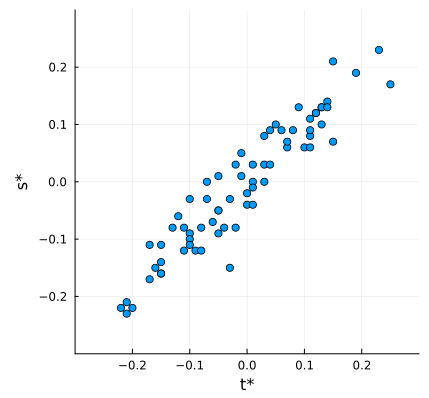}
\centering
\caption{{\bf Optimal Projection Time}. $t^*$ vs $s^*$ comparison that uses the canonical pair to map to
  the original data. $t^*$ and $s^*$ show a linear relationship between one
  another. Thus a simple linear regression model would be suitable to
  map $t^*$ to $s^*$.}
  
\label{fig:s_vs_t}
\end{figure}

Given that the two sets of data are assumed to be associated, the $(t^*,s^*)$ derived from these data sets is expected to preserve some level of association with one another. A fortiori, we empirically found in our experiments (cf., Section \ref{sec:Experiments}) that the dependency between the optimal $(t^{*}_i,s^{*}_i)$ is even linear, and even close to the identity, as shown at Figure \ref{fig:s_vs_t}.  We have not yet established a stringent theoretical justification for this phenomenon, and whether this is a general property of the algorithm, a consequence of the construction of $t$ and $s$ from normalized vectors, it stems from constraining ourselves the Lie manifold property, or a peculiarity of the particular experimental scenario. 


We proceed by recursively defining the next ICCA pairs. 

\subsubsection{Next ICCA Pairs}\label{sec:nextpairs} Denote the projection of
$\mu_x^{-1}x_i$ on $H_{{v}^{(1)}}$ by $\mbox{Proj}^{(1)}(x_i)$,
and the projection of $\mu_y^{-1}y_i$ on $H_{{u}^{(1)}}$ by
$\mbox{Proj}^{(1)}(y_i)$. 

Firstly, we remove $\mbox{Proj}^{(1)}(x_i)$ from $\mu_x^{-1}x_i$ and $\mbox{Proj}^{(1)}(y_i)$ from $\mu_y^{-1}y_i$, which results in $x_i^{(2)}$ and $y_i^{(2)}$, respectively. Then, given $\forall i\;:\:(x_i^{(1)},y_i^{(1)}) = (\exp(t^*_iv^{(1)}),\exp(s^*_iu^{(1)}))$, we define the second ICCA pair by:
\begin{align}
{v}^{(2)}, {u}^{(2)} = \underset{\substack{||v||=||u||=1}}{\mbox{argmin}}\sum_{i=1}^N \Bigl(&D^2(x_i^{(2)}, H_{{v}})\label{eq:2stPair}\\
+&D^2(y_i^{(2)}, H_{{u}})\nonumber \\
+&D^2\bigl(\mbox{Proj}_{H_{{v}}}(x_i^{(2)}), \mbox{Proj}_{H_{{u}}}(y_i^{(2)})\bigr) 
\Bigr).\nonumber 
\end{align}
In general, the $(k+1)$-th ICCA pair is recursively defined by:
\begin{align}
{v}^{(k+1)}, {u}^{(k+1)} = \underset{\substack{||v||=||u||=1}}{\mbox{argmin}}\sum_{i=1}^N \Bigl(&D^2(x_i^{(k)}, H_{{v}})\label{eq:kthPair}\\
+&D^2(y_i^{(k)}, H_{{u}})\nonumber \\
+&D^2\bigl(\mbox{Proj}_{H_{{v}}}(x_i^{(k)}), \mbox{Proj}_{H_{{u}}}(y_i^{(k)})\bigr) 
\Bigr).\nonumber 
\end{align}
where $x^{(k)}$ and  $y^{(k)}$ are the the residual data pair from the last iteration.
The  solution to the ICCA problem consists of all $k$ ICCA pairs 
$\bigl({v}^{(k)}, {u}^{(k)}\bigr)$ and their corresponding mappings $f^{(k)}(t)$ between the optimal projection times. 

The equations Eq.\eqref{eq:2stPair} and Eq.\eqref{eq:kthPair} represent the  decomposition of data into the $k$ canonical pairs. We present two algorithms to decompose and reconstruct two data sets on the Lie Manifold using the first canonical pair Eq.\eqref{eq:1stPair_eplicite}, which can be  extended to the $k$-th canonical pair by Eq.\eqref{eq:2stPair} and Eq.\eqref{eq:kthPair}. We show in Section \ref{sec:Experiments} that the first ICCA pair  results in a significantly lower prediction error in comparison to the standard Euclidean CCA.

\begin{enumerate}
    \item ICCA Decomposition: extracts the first canonical pair, and calculates the model, $\hat{s}$, mapping  one optimal projection time, $t^*$, to the other, $s^*$. The subsequent pairs can be calculated by applying Eq.~\eqref{eq:kthPair}.
    \item ICCA Reconstruction: Predicts $y\in G$ from $x\in G$ using the first canonical pair.
\end{enumerate}


 \subsection{ICCA Decomposition}

  The ICCA decomposition uses iteration to solve for the minimization for the first canonical pair. The algorithm contains two stages: initialization and iteration.

{The distance, $D(\cdot, \cdot)$ between a point on the manifold and a one-dimensional sub-manifold has multiple local minima. We initialize the algorithm by firstly finding the optimal projection of the data set to the first canonical pair ($v^{(1)},u^{(1)}$), lines 2-4 in Alg.~\ref{TwoStepICCA}. Then, we optimize the full objective in Eq.\eqref{eq:1stPair_eplicite} until convergence, lines 5-9 in Alg.\ref{TwoStepICCA}. Note that, for this, the iteration stage alternates between finding the   optimal canonical pair $(v^{(1)},u^{(1)})$ and its respective optimal projection times for the data, $\{t^*_i, s^*_i\}_{i=1}^N$. 
}

At convergence, we estimate the predictor for $s^*$ from $t^*$ using linear regression, Eq.\eqref{eq:OptimalProjectionTime}. The algorithm returns the model $\hat{s}$, and the optimal first canonical pair $(v^{(1)},u^{(1)})$.


\begin{algorithm}[t]
			\begin{algorithmic}[1] 
				\STATE {\bf Input:} $\{x_i,y_i\}_{i=1}^N \in G$ - data pairs on the manifold, $G$.

                    \STATE \bf Initialize
				\STATE ${v},t^* \leftarrow \underset{||{v}||=1}{ {\mbox{argmax}  }} \;\sum_{i=1}^N\underset{t}{ \mbox{ min }}\;||\log\bigl(\mu_x^{-1}x_i\exp(t{v})\bigr)||^2,$
				\STATE ${u},s^* \leftarrow \underset{||{u}||=1}{ \mbox{ argmax  }} \;\sum_{i=1}^N\underset{s}{ \mbox{ min }}\;||\log\bigl(\mu_y^{-1}y_i\exp(s{u})\bigr)||^2,$
			\REPEAT
				\STATE ${v},{u} \leftarrow {\mbox{Eq.}} \eqref{eq:1stPair_eplicite}$\hfill\COMMENT{{\normalfont with fixed} $\{t^*_i,s^*_i\}$}
    
                    \STATE $\{t^*_i\}^N_{i=1} \leftarrow  \bigl\{\underset{t}{\mbox{ argmin}} \; D(\mu_x^{-1}x_i, \exp(s{v}))\bigr\}_{i=1}^N$

                    \STATE $\{s^*_i\}^N_{i=1} \leftarrow  \bigl\{\underset{s}{\mbox{ argmin}} \; D(\mu_y^{-1}y_i, \exp(t{u}))\bigr\}_{i=1}^N$
				\UNTIL{convergence}
                    \STATE $\hat{s}(\cdot)\leftarrow \mbox{Eq.} \eqref{eq:OptimalProjectionTime}$
                    
				\RETURN{ $\{{v}^{(1)},{u}^{(1)}\}$ and $\hat{s}(\cdot)$.} 
			\end{algorithmic} 
   \caption{ICCA Decomposition} \label{TwoStepICCA} 
		\end{algorithm}
  
\subsection{ICCA Reconstruction} \label{sec:ICCAReconstruction}




Given the $k$ ICCA pairs $({v}^{(k)}, {u}^{(k)})$ and the model for the optimal prediction time, $\hat{s}(t)$, we can use these  to reconstruct a secondary point $\hat{y}(x)\in {G}$ on the Lie manifold from a primary point $x\in {G}$ on the Lie manifold, as follows: 
\begin{align}
 &t^{(1)}  = \underset{t}{\mbox{ argmin}} \; D(x, \exp(t{v}^{(1)}))\nonumber\\
  &\forall k\ge 1\;:\;  t^{(k+1)}  = \underset{t}{\mbox{ argmin}} \;  D(x^{(k)}, \exp(t{v}^{(k)}))\nonumber\\
&\hat{y} = \exp\bigl(\sum_{k=1}^K\hat{s}^{(k)}(t^{(k)} ){u}^{(k)}\bigr)\label{eq:recoYfull}\in {G},
 \end{align}
 where $x^{(k)}$ is as explained after Eq. \eqref{eq:kthPair} in Section \ref{sec:nextpairs}.

\section{Experiments}\label{sec:Experiments}

The state space in articulated robotic systems \cite{LieRobotics} lies within a manifold consisting of group elements such
as rotations and translations. The novelty of ICCA is the ability to account for these intrinsic properties of the robotic states. A more consistent representation of such states than provided by standard CCA would be one that remains inside the manifold structure, leading to improved CCA output accuracy. The goal of the experiments is to answer the following questions: 
\begin{enumerate}
\item Can a solution be effectively calculated from a finite sample of point pairs on the Lie manifold? 
\item Is the ICCA reconstruction more accurate than that achieved by the standard Euclidean CCA? 
\end{enumerate}

\subsection{Experimental settings}
The MuJoCo simulator \cite{todorov2012mujoco} was used to generate the data for the evaluation of the ICCA method on the anthropomorphic hand. The hand consisted of $14$ finger joints; the inner finger joint was represented as a 3D Special Orthogonal SO(3) group, and the other finger joints as a 2D Special Orthogonal SO(2) group. 

\begin{figure}[t!]

\includegraphics[scale=0.21]{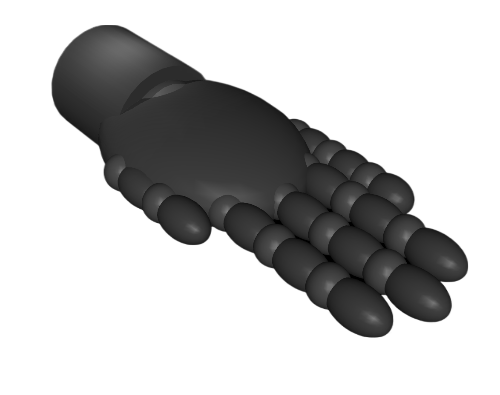}
\includegraphics[scale=0.21]{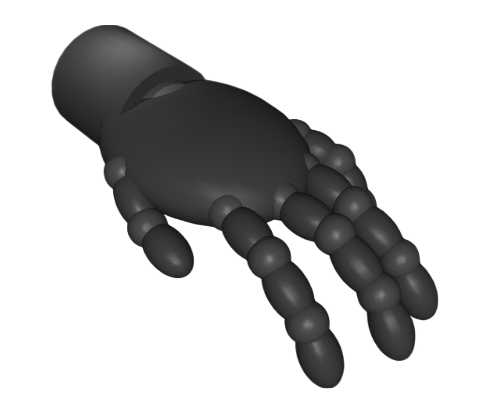}
\includegraphics[scale=0.21]{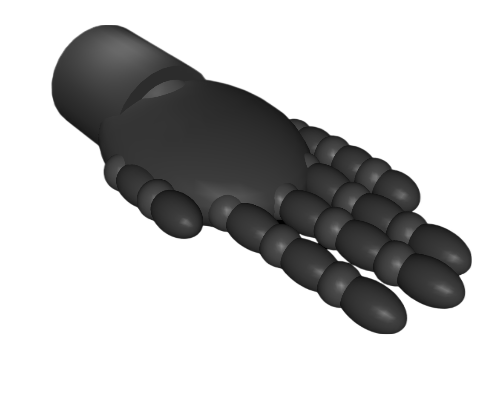}
\includegraphics[scale=0.21]{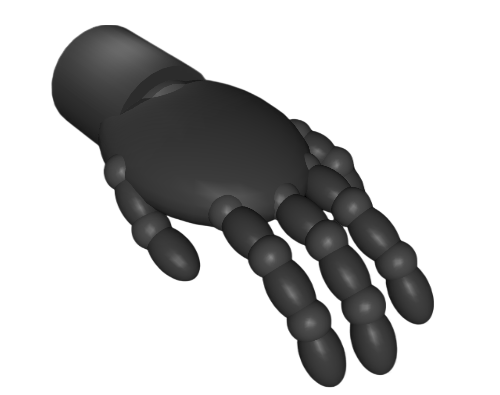}
\centering
\caption{Anthropomorphic Robotic Hand used in Sec.\ref{sec:Experiments}. The images above show one of the simulations along with its ICCA reconstruction. Top-left represents the original state. Bottom-left represents the initial configuration (the neutral state with noise). Top-right represents the ground-truth configuration. Bottom-right represents the ICCA reconstruction given the initial configuration.}
\label{fig:MuJoCoHand}
\end{figure}

The two data sets for the ICCA analysis consist of the original configurations of the hand and the final configurations after $20$ simulation steps with a stochastic action sequence 'Action Noise', as explained below. An example of the original (current) configuration, $x\in G$, is shown in the top left image at Figure \ref{fig:MuJoCoHand}, and an example for the final configuration, $y\in G$, is shown at the top right image. 

The set of the current configurations, $\{x_i\}_{i=1}^N\in G$, is generated by adding a stochastic perturbation (noise) to $x\in G$, as explained below. Each of the final configurations, $\{y_i\}_{i=1}^N\in G$, corresponds to a particular $\{x_i\} \in G$ after applying the predefined action sequence with small perturbation in each action. An example of a perturbed current configuration and the corresponding final configuration of the hand are shown at the bottom left and right images of Figure \ref{fig:MuJoCoHand}, respectively.

To create the original configuration set $\{x_i\}_{i=1}^N\in G$, we applied noise with the following features:

\begin{enumerate}
\item {\bf Configuration Noise:} Gaussian noise with zero mean vector with dimensionality $14$ and diagonal covariance matrix, $\Sigma_{14\time 14}=0.02\,\mathbf{I}_{14\times 14}$. This noise is added to the original configuration.
\item {\bf Action Noise:} Gaussian noise with zero mean, $\mu=0.0$ and variance, $\sigma=0.01$. This noise is independently added to each action in the predefined action sequence with length $20$.
\end{enumerate}

The total number of data points, $\{x_i, y_i\}_{i=1}^{N}$, is $N=10\times 1500$, split into 10 experiments with $1500$ data points in each experiment. $2:1$ 'train-to-test' split was performed on the data. We used the training data set for the calculations of the first ICCA pair, $(u^{(1)}, v^{(1)})$, and of the regression model $\hat{s}(t)$. The test set is used for measuring the reconstruction error between the true final configuration, $y$, in $(x, y)$, and the reconstructed configuration, $\hat{y}(x)$, with $(u^{(1)}, v^{(1)})$ and $\hat{s}(t)$, as explained in Section \ref{sec:ICCAReconstruction}. The reconstruction error was defined as the Mean Squared Error (MSE) between $\hat{y} = \hat{y}(x)$ and $y$. We compared the reconstruction error between the conventional CCA (based on the Euclidean distance) and the proposed ICCA (based on intrinsic distance).

\subsection{Results}

\begin{figure}[t!]
\includegraphics[scale=0.35]{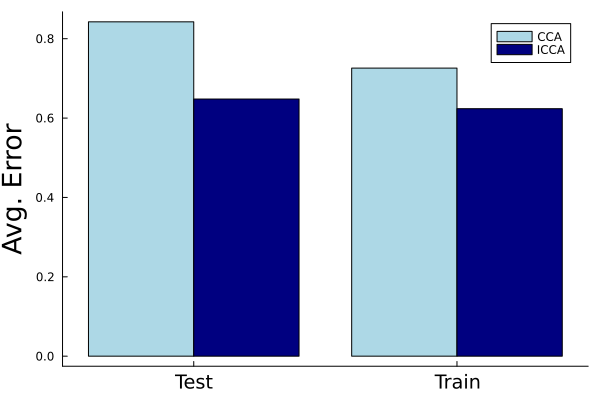}
\centering
\caption{MSE between the true and the reconstructed configurations of the anthropomorphic robotic arm for CCA and ICCA in training and testing.}
\label{fig:barchart}
\end{figure}

Accuracy and generalization are improved significantly for ICCA compared to CCA. The MSE comparison shown at Figure \ref{fig:barchart}. The training improvement achieved $16.41\%$, while the testing improvement was $23.08\%$. ICCA achieved better generalization as the train-test accuracy difference was $3.77\%$ for ICCA and $16.05\%$ for CCA.






There is an initial loss decrease in the initialization stage, (lines 2-4, Alg.\eqref{TwoStepICCA}), where the initial values for $v^{(1)}$ and $u^{(1)}$ are found. Then, they are fine-tuned in the iteration stage (lines 5-9 Alg. \ref{TwoStepICCA}) until convergence (Figure \ref{fig:loss convergance}). The plot shows the average loss over ten experiments with a very small variance.




\addtolength{\textheight}{-3cm}   
{


\section{Conclusions and Future Work}\label{sec:conclusion}
This work presents a novel method to generalize CCA to the nonlinear setting
of a Lie manifold. The distance optimality, projection criteria, and subspace concepts generalize naturally to the Lie setting via intrinsic Riemannian distances and geodesics, respectively. This setting is a central application in the context of articulated robotic devices \cite{Hardoon2004CCA}.

The projection-based approach opens doors to symmetry-aware methodologies, among these, the learning of parameters in a transformation model.


Our formalism expressly uses the group-theoretic properties of the Lie manifold rather than merely approximating the Lie manifold, e.g.\ via kernel-based approaches. However, further refinements are based on explicitly symmetry-respecting learning methods, such as kernels \cite{kondor08:_group_theor_method_machin_learn}. With the presented generalization, the option to incorporate further direct tools from the theory of groups has now become available to enhance the quality of treatment of systems with intrinsically symmetric structures.



{We also emphasize that, like the whole family of PCA, CCA, and their informational generalization, the  Information Bottleneck methods, the ICCA method enables a controlled hierarchical dimensional reduction of dynamical control systems respecting the constraint manifold and thus allows us to control complexity without losing the structural guarantees enforced by the constraints, and thus to substantially limit the performance loss due to the approximation.}

We defer to future work the theoretical study of the relationship between the optimal projection times, which was found to be linear in the Lie manifold and the extension of the introduced ICCA to the data on different manifolds which are not necessarily Lie manifolds.


\begin{figure}[t!]
\includegraphics[scale=0.35]{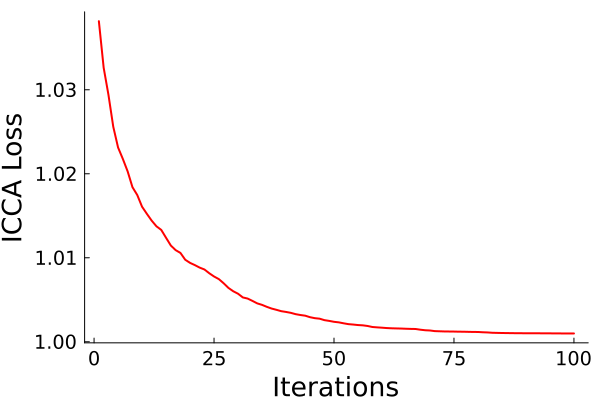}
\centering
\caption{Convergence of the ICCA loss for Grasping Hand in the iteration stage. The loss plot is created by taking the average on $10$ sets of experiments on the grasping hand task. The curve is based on loss in the iteration stage of ICCA Decomposition. There is an additional loss decrease in the initialization stage.}
\label{fig:loss convergance}
\end{figure}

}

{
\section*{ACKNOWLEDGMENTS}

The authors gratefully acknowledge the contribution of NSF and the support of the
Pazy Foundation under ID 195. The authors thank the graduate students at the CI$^2$ Lab, CoE, SJSU, Ezgi Kaya, Paul Mello and Alex Zaliznyak for proofreading the manuscript.
}


{








\bibliographystyle{IEEEtran}
\bibliography{ref}

\begin{thebibliography}{10}
\providecommand{\url}[1]{#1}
\csname url@samestyle\endcsname
\providecommand{\newblock}{\relax}
\providecommand{\bibinfo}[2]{#2}
\providecommand{\BIBentrySTDinterwordspacing}{\spaceskip=0pt\relax}
\providecommand{\BIBentryALTinterwordstretchfactor}{4}
\providecommand{\BIBentryALTinterwordspacing}{\spaceskip=\fontdimen2\font plus
\BIBentryALTinterwordstretchfactor\fontdimen3\font minus
  \fontdimen4\font\relax}
\providecommand{\BIBforeignlanguage}[2]{{%
\expandafter\ifx\csname l@#1\endcsname\relax
\typeout{** WARNING: IEEEtran.bst: No hyphenation pattern has been}%
\typeout{** loaded for the language `#1'. Using the pattern for}%
\typeout{** the default language instead.}%
\else
\language=\csname l@#1\endcsname
\fi
#2}}
\providecommand{\BIBdecl}{\relax}
\BIBdecl

\bibitem{tishby2000information}
N.~Tishby, F.~C. Pereira, and W.~Bialek, ``The information bottleneck method,''
  \emph{The 37th annual Allerton Conference on Communication, Control, and
  Computing}, 2000.

\bibitem{slonim2002information}
N.~Slonim, ``The information bottleneck: Theory and applications,'' Ph.D.
  dissertation, Hebrew University of Jerusalem, Israel, 2002.

\bibitem{katayama2005subspace}
T.~Katayama \emph{et~al.}, \emph{Subspace methods for system
  identification}.\hskip 1em plus 0.5em minus 0.4em\relax Communications and
  Control Engineering, Springer, 2005, vol.~1.

\bibitem{chechik2003information}
G.~Chechik, A.~Globerson, N.~Tishby, and Y.~Weiss, ``Information bottleneck for
  gaussian variables,'' \emph{Advances in Neural Information Processing
  Systems}, vol.~16, 2003.

\bibitem{creutzig2009past}
F.~Creutzig, A.~Globerson, and N.~Tishby, ``Past-future information bottleneck
  in dynamical systems,'' \emph{Physical Review E}, vol.~79, no.~4, p. 041925,
  2009.

\bibitem{amir2015past}
N.~Amir, S.~Tiomkin, and N.~Tishby, ``Past-future information bottleneck for
  linear feedback systems,'' in \emph{2015 54th IEEE Conference on Decision and
  Control (CDC)}.\hskip 1em plus 0.5em minus 0.4em\relax IEEE, 2015, pp.
  5737--5742.

\bibitem{klami2013bayesian}
A.~Klami, S.~Virtanen, and S.~Kaski, ``Bayesian canonical correlation
  analysis.'' \emph{Journal of Machine Learning Research}, vol.~14, no.~4,
  2013.

\bibitem{andrew2013deep}
G.~Andrew, R.~Arora, J.~Bilmes, and K.~Livescu, ``Deep canonical correlation
  analysis,'' in \emph{International conference on machine learning}.\hskip 1em
  plus 0.5em minus 0.4em\relax PMLR, 2013, pp. 1247--1255.

\bibitem{wang2015stochastic}
W.~Wang, R.~Arora, K.~Livescu, and N.~Srebro, ``Stochastic optimization for
  deep cca via nonlinear orthogonal iterations,'' in \emph{2015 53rd Annual
  Allerton Conference on Communication, Control, and Computing
  (Allerton)}.\hskip 1em plus 0.5em minus 0.4em\relax IEEE, 2015, pp. 688--695.

\bibitem{88036}
M.~Cutkosky and I.~Kao, ``Computing and controlling compliance of a robotic
  hand,'' \emph{IEEE Transactions on Robotics and Automation}, vol.~5, no.~2,
  pp. 151--165, 1989.

\bibitem{Ahmed2015Sensor}
A.~A. et~al., ``Pressure sensor: State of the art, design, and application for
  robotic hand,'' \emph{Journal of Sensors}, 2015.

\bibitem{5460699}
J.~L. Raheja, R.~Shyam, U.~Kumar, and P.~B. Prasad, ``Real-time robotic hand
  control using hand gestures,'' in \emph{2010 Second International Conference
  on Machine Learning and Computing}, 2010, pp. 12--16.

\bibitem{fletcher2003gaussian}
P.~T. Fletcher, S.~Joshi, C.~Lu, and S.~M. Pizer, ``Gaussian distributions on
  lie groups and their application to statistical shape analysis,'' in
  \emph{Information Processing in Medical Imaging: 18th International
  Conference, IPMI 2003, Ambleside, UK, July 20-25, 2003. Proceedings
  18}.\hskip 1em plus 0.5em minus 0.4em\relax Springer, 2003, pp. 450--462.

\bibitem{moakher2002means}
M.~Moakher, ``Means and averaging in the group of rotations,'' \emph{SIAM
  journal on matrix analysis and applications}, vol.~24, no.~1, 2002.

\bibitem{bonfiglioli2011topics}
A.~Bonfiglioli and R.~Fulci, \emph{Topics in noncommutative algebra: the
  theorem of Campbell, Baker, Hausdorff and Dynkin}.\hskip 1em plus 0.5em minus
  0.4em\relax Springer Science \& Business Media, 2011, vol. 2034.

\bibitem{agrachev2013control}
A.~A. Agrachev and Y.~Sachkov, \emph{Control theory from the geometric
  viewpoint}.\hskip 1em plus 0.5em minus 0.4em\relax Springer Science \&
  Business Media, 2013, vol.~87.

\bibitem{fletcher2003statistics}
P.~Fletcher, C.~Lu, and S.~Joshi, ``Statistics of shape via principal component
  analysis on lie group,'' in \emph{Proceedings of CVPR}, 2003.

\bibitem{PCALie}
H.~P. Jameson~Cahill, Dustin G.~Mixon, ``Lie pca: Density estimation for
  symmetric manifolds,'' in \emph{sciencedirect}, 2023.

\bibitem{Kim2014Images}
T.~K. Kim, S.~F. Wong, and R., ``Tensor canonical correlation analysis for
  action classification,'' in \emph{In Computer Vision and Pattern
  Recognition}.\hskip 1em plus 0.5em minus 0.4em\relax IEEE, 2014, pp.
  529--545.

\bibitem{Arora2012Sound}
R.~Arora and K.~Livescu, ``Kernel cca for multi-view learning of acoustic
  features using articulatory measurements,'' in \emph{In Proceedings of the
  Symposium on Machine Learning in Speech and Language Processing}.\hskip 1em
  plus 0.5em minus 0.4em\relax ISCA, 2012, pp. 33--37.

\bibitem{Lisanti2014Camera}
G.~Lisant, I.~Masi, and A.~Bimbo, ``Matching people across camera views using
  kernel canonical correlation analysis,'' in \emph{Proceedings of the
  International Conference on Distributed Smart Cameras}.\hskip 1em plus 0.5em
  minus 0.4em\relax ACM, 2014, pp. 1--6.

\bibitem{Chen2021Robot}
L.~Chen, P.~Chen, S.~Zhao, Z.~Luo, W.~Chen, Y.~Pei, H.~Zhao, J.~Jiang, M.~Xu,
  Y.~Yan, and E.~Yin, ``Adaptive asynchronous control system of robotic arm
  based on augmented reality-assisted brain-computer interface,'' \emph{Journal
  of Neural Engineering}, vol.~18, no.~6, 2021.

\bibitem{LieRobotics}
J.~Selig, \emph{Lie Groups and Lie Algebras in Robotics}.\hskip 1em plus 0.5em
  minus 0.4em\relax Computational Noncommutative Algebra and Applications,
  2004.

\bibitem{todorov2012mujoco}
E.~Todorov, T.~Erez, and Y.~Tassa, ``Mujoco: A physics engine for model-based
  control,'' in \emph{2012 IEEE/RSJ International Conference on Intelligent
  Robots and Systems}.\hskip 1em plus 0.5em minus 0.4em\relax IEEE, 2012, pp.
  5026--5033.

\bibitem{Hardoon2004CCA}
D.~R. Hardoon, S.~Szedmak, and J.~Shawe-Taylor, ``Canonical correlation
  analysis: An overview with application to learning methods,'' in \emph{Neural
  Computation}.\hskip 1em plus 0.5em minus 0.4em\relax IEEE, 2004, pp.
  2639--2664.

\bibitem{kondor08:_group_theor_method_machin_learn}
R.~Kondor, ``Group theoretical methods in machine learning,'' Ph.D.
  dissertation, Columbia University, 2008.

\bibitem{howe1983very}
R.~Howe, ``Very basic lie theory,'' \emph{The American Mathematical Monthly},
  vol.~90, no.~9, pp. 600--623, 1983.

\bibitem{hall2003lie}
\BIBentryALTinterwordspacing
B.~Hall and B.~Hall, \emph{Lie Groups, Lie Algebras, and Representations: An
  Elementary Introduction}, ser. Graduate Texts in Mathematics.\hskip 1em plus
  0.5em minus 0.4em\relax Springer, 2003. [Online]. Available:
  \url{https://books.google.com/books?id=m1VQi8HmEwcC}
\BIBentrySTDinterwordspacing

\bibitem{selig2004lie}
J.~M. Selig, ``Lie groups and lie algebras in robotics,'' in
  \emph{Computational Noncommutative Algebra and Applications}.\hskip 1em plus
  0.5em minus 0.4em\relax Springer, 2004.

\bibitem{bourmaud2015continuous}
G.~Bourmaud, R.~M{\'e}gret, M.~Arnaudon, and A.~Giremus, ``Continuous-discrete
  extended kalman filter on matrix lie groups using concentrated gaussian
  distributions,'' \emph{Journal of Mathematical Imaging and Vision}, vol.~51,
  no.~1, pp. 209--228, 2015.

\bibitem{featherstone2014rigid}
R.~Featherstone, \emph{Rigid body dynamics algorithms}.\hskip 1em plus 0.5em
  minus 0.4em\relax Springer, 2014.

\bibitem{sola2018micro}
J.~Sola, J.~Deray, and D.~Atchuthan, ``A micro lie theory for state estimation
  in robotics,'' \emph{arXiv preprint arXiv:1812.01537}, 2018.

\end{thebibliography}
\setcounter{section}{0}
\section{APPENDIX: Elements of Lie group theory}\label{sec1:firstappendix}

{\it A finite-dimensional Lie group} is a group that is at the same time a differentiable manifold  \cite{howe1983very,hall2003lie}. For every point $x\in G$ on the manifold, there exists a
tangent linear vector space ${TG}_{x}$. The tangent space at
the identity element,
$TG_{e}$, 
is special. It is called the \emph{Lie algebra}, $\mathfrak{g}$, of
the Lie group, ${G}$.
One can map elements of the Lie group (manifold) to those of its
algebra (linear vector space with an associative bilinear product) and vice versa by:
\begin{align}
  \label{eq:connection}
\mathfrak{g} \stackrel[\exp]{\log}\longleftrightarrows {G}
\end{align}
where '$\exp$' and '$\log$' are calculated via the corresponding Taylor
series 
of the operators. Concretely,
$\forall {g}\in \mathfrak{g}$, $\exp({g})
=\sum_{n=0}^{\infty}\frac{{g}^{n}}{n!}$, where ${g}^{n}{=}{{g}\circ
  {g}\circ\dots\circ  {g}}$ is the $n$-fold product of ${g}$ in the Lie
algebra and for the sum to be well-defined, we interpret the
expression in terms of
the matrix representation of the operators, which operate on $k-$dimensional linear vector space of the Lie algebra. 

{\it The Lie algebra}, as a linear vector space, is spanned by a basis of
$k$ elements, $E=\{E_1, E_2, \dots, E_k\}$, where $k$ is the dimension 
of the manifold $G$ and $E_i\in \mathbb{R}^k$. Thus every element  in the algebra, ${g} \in
\mathfrak{g}$, interpreted as $k$-dimensional vector, is represented by a unique linear combination of the basis elements 
\begin{align}\label{eq:algebraelement}
    {g}({\small\alpha}) = \sum_{i=1}^k {\small\alpha_i} E_i,\mbox{ with } k \mbox{ scalars }, {\alpha} = \{{\small\alpha}_i\}_{i=1}^k.
\end{align}
Eq.~\eqref{eq:connection} and
\eqref{eq:algebraelement} induce a mapping of the collection ${\alpha}$  to the Lie 
group, $G$, as follows: $G(\alpha) = \exp({g}(\alpha)) \in {G}$.

The formalism is of particular interest for the special case of articulated robotic systems, since its configuration space (under multiple concatenated links) is represented by the Lie groups of rotations and translations, reviewed below. 

\subsubsection{Groups of rotation and translation}
The Lie groups of 2D/3D spatial rotations, $SO(2)$/$SO(3)$, and
translations, $SE(2)$/$SE(3)$, fully characterize the primitive
geometric motions of rigid bodies, and are widely used in robotics
\cite{selig2004lie,bourmaud2015continuous,featherstone2014rigid}. Both
groups admit matrix representations \cite{hall2003lie}. The group composition operator, ${\circ}$, is
 the standard matrix multiplication. and $\exp$ with the matrix exponential.  


The exponential map from the $so(3)$ algebra to the
corresponding $SO(3)$ Lie group in Eq.~\eqref{eq:connection} can be explicitly
calculated by the Rodrigues rotation formula
\cite{sola2018micro}:
\begin{align}
R = \exp\bigl(\theta U\bigr) = {I} + U\sin \theta + U^2(1- \cos \theta)\in \mathbb{R}^{3\times 3}\label{eq:expmap},
\end{align}
where ${I}$ is the identity matrix, $\theta\in\mathbb{R}$ the angle of rotation and $U$ the matrix (the element of the Lie algebra) generating the rotation. The Rodrigues rotation formula \cite{sola2018micro} 
\begin{align}
 \log (R) = \frac{\theta(R-R^T)}{2\sin \theta}\label{eq:logmap}, \qquad \theta = \cos^{-1}\Biggl(\frac{\mbox{trace}(R)-1}{2}\Biggr)\nonumber
\end{align}
together with the 'logarithmic map' allows us to effectively calculate the intrinsic distance and the projections to subgroups, which are the important components of the proposed ICCA method.

}

\end{document}